# 基于局部回归融合的多核聚类方法

杜 亮[1,2] 任 鑫[1] 张海莹[1] 周 芃[3]
1 山西大学计算机与信息技术学院 太原 030006
2 山西大学大数据科学与产业研究院 太原 030006
3 安徽大学计算机科学与技术学院 合肥 230601
(im.duliang@qq.com)

**摘 要** 针对现有多核聚类方法较少考虑多核数据局部流形结构以及在多核融合时学习参数过多进而易受多核噪声异常等干扰的问题,文中首先提出了基于局部核回归的聚类方法(CKLR)。该方法通过局部学习来刻画单核数据的流形结构并采用稀疏化的局部核回归系数来进行预测和聚类。文中进一步提出了基于单核局部核回归融合的多核聚类方法(CMKLR)。该方法为每个核矩阵构造对应的稀疏化的局部核回归系数,并采用全局线性加权融合的方式获得了多核数据下的局部流形结构和同样稀疏化的多核局部回归系数。所提方法较好地避免了现有方法的两个缺陷,且该方法仅包含局部邻域大小这一超参数。实验结果表明,所提方法在测试数据集上的聚类性能优于当前的主流多核聚类方法。

# Multiple Kernel Clustering via Local Regression Integration

DU Liang[1,2], REN Xin[1], ZHANG Hai-ying[1] and ZHOU Peng[3]
1 School of Computer and Information Technology, Shanxi University, Taiyuan 030006, China
2 Institute of Big Data Science and Industry, Shanxi University, Taiyuan 030006, China
3 School of Computer Science and Technology, Anhui University, Hefei 230601, China

**Abstract** Multiple kernel methods less consider the intrinsic manifold structure of multiple kernel data and estimate the consensus kernel matrix with quadratic number of variables, which makes it vulnerable to the noise and outliers within multiple candidate kernels. This paper first presents the clustering method via kernelized local regression (CKLR). It captures the local structure of kernel data and employs kernel regression on the local region to predict the clustering results. Moreover, this paper further extends it to perform clustering via the multiple kernel local regression (CMKLR). We construct the kernel level local regression sparse coefficient matrix for each candidate kernel, which well characterize the kernel level manifold structure. We then aggregate all the kernel level local regression coefficients via linear weights and generate the consensus sparse local regression coefficient, which largely reduces the number of candidate variables and becomes more robust against noises and outliers within multiple kernel data. Thus, the proposed method CMKLR avoids the above two limitations. It only contains one additional hyper parameter for turning. Extensive experimental results show that the clustering performance of the proposed method on benchmark data set is better than that of 10 state-of-the-art multiple kernel clustering methods.

## 1 引言

聚类是数据挖掘和机器学习的核心问题之一。核方法通过核函数将原始数据从线性空间变换到非线性空间,从而使线性难分的数据在核空间中变得易分[1]。核方法已经被广泛用于多种学习任务,如基于核方法的分类[2]、聚类和最优实验设置[3]等。核方法用于聚类时面临的问题之一是如何根据不同类型的数据集选择和设计合适的核函数及其相关的超参数。多核聚类通过联合建模学习,一方面利用多核数据进行聚类,另一方面利用聚类结果指导多核数据的融合。这样既减轻了核函数选择的困难程度,又提升了聚类精度和稳定性,因而多核聚类成为了当前国内外研究的热点。

代表性多核聚类方法主要包括多核 $K$-means 方法[4]、多核谱聚类方法[5-10]和多核子空间聚类方法[11-13]。Du 等采用 $L_{21}$ 损失提出鲁棒多核 $K$-means 方法(Robust Multiple Kernel $K$-Means using L2,1-Norm, RMKKM)[4];Liu 等提出核关系诱导的矩阵正则化的多核 $K$-means 方法(Multiple Kernel $K$-Means Clustering with Matrix-induced Regularization,

到稿日期:2020-10-18 返修日期:2021-01-29 本文已加入开放科学计划(OSID),请扫描上方二维码获取补充信息。
基金项目:国家自然科学基金(61976129,61806003)
This work was supported by the National Natural Science Foundation of China(61976129,61806003).
通信作者:周芃(zhoupeng@ahu.edu.cn)

MKKMMR)[8]；Li 等提出基于局部对齐准则的多核聚类方法(Multiple Kernel Clustering with Local Kernel Alignment Maximization，LKAMKC)[9]；Liu 等进一步提出了基于最优邻近核学习的多核聚类方法(Optimal Neighborhood Kernel Clustering with Multiple Kernels，ONMKC)[10]；Kang 等提出基于低秩学习的多核聚类方法[11]；Yang 等提出基于自适应鲁棒多核子空间学习的聚类方法(Joint Correntropy Metric Weighting and Block Diagonal Regularizer for Robust Multiple Kernel Subspace Clustering，JMKSC)[12]；Zhan 等提出基于一致性结构化图学习的多视图聚类方法(Multiview Consensus Graph Clustering，MCGC)[13]。近年来，研究发现，与单核单视图聚类相比，多核聚类面临更加严重的数据质量问题，即数据缺失、噪声异常等。为此，研究人员提出了多种鲁棒多核聚类方法[14]和面向缺失场景的不完备多核聚类方法[15]。

上述现有多核聚类方法常通过线性加权的策略来构建或者近似最优核，这类方法存在两方面的问题：1)尽管核函数较好地刻画了数据的非线性关系[16]，但是，正如 Cai 等[17]所指出的，核函数构造的核矩阵并没有很好地反映数据间的局部流形结构；2)进行最优核学习时，既需要学习所有样本对两两之间的关系，又需要满足半正定等基本要求。此时融合核容易受少量噪声异常候选核的干扰，进而降低最优核质量。

为了一定程度地解决上述两个问题，本文提出了基于局部回归融合的多核聚类方法(Multiple Kernel Clustering via Local Regression Integration，CMKLR)。具体来讲，该方法首先为某样本在某个核函数下筛选其对应的邻近样本集，并基于该样本集和该核函数构造局部回归模型，进而借助该模型对该样本的聚类标签进行预测。为了获得更加精准和稳定的预测结果，我们采用线性加权的方式将该样本下的多个局部回归模型融合得到线性混合的多核局部回归模型。我们将所有样本的混合局部回归模型的预测误差进行建模，最终得到基于局部回归的多核聚类方法。值得指出的是，该方法一方面在单核意义下考虑局部流形结构，并以多核线性融合的方式在多核意义下考虑数据内在的局部流形结构，另一方面仅仅构造和学习多个局部稀疏化的回归系数，避免了传统方法对核矩阵所有参数的学习，极大地缩减了待求解参数的规模，进而减小了低质量数据对聚类任务的影响。基准数据集上的实验结果充分表明，该方法在多种不同类型的数据上都明显优于当前主流的方法。

## 2 基于局部回归融合的多核聚类方法

给定数据集 $\boldsymbol{X}=\{\boldsymbol{x}_i\}_{i=1}^n$，其中$\boldsymbol{x}_i\in\mathbb{R}^{d\times 1}$，聚类任务的目标是将样本划分到 $c$ 个簇中$\{C_j\}_{j=1}^c$。给定核函数 $k(x,y)\rightarrow\mathbb{R}$，根据核技巧得到其对应的核矩阵 $\boldsymbol{K}\in\mathbb{R}^{n\times n}$。假设一共给定 $m$ 个核函数，我们共获得 $m$ 个核矩阵$\{\boldsymbol{K}^r\}_{r=1}^m$用于聚类。定义划分矩阵 $\boldsymbol{P}\in\mathbb{R}^{n\times c}$，其中如果样本 $\boldsymbol{x}_i$属于簇 $C_j$，则记 $\boldsymbol{P}_{ij}=1$，否则记为 $\boldsymbol{P}_{ij}=0$。定义缩放划分矩阵 $\boldsymbol{Y}\in\mathbb{R}^{n\times c}$，其中 $\boldsymbol{Y}=\boldsymbol{P}(\boldsymbol{P}^{\mathrm{T}}\boldsymbol{P})^{-\frac{1}{2}}$，容易验证 $\boldsymbol{Y}^{\mathrm{T}}\boldsymbol{Y}=\boldsymbol{I}$。

### 2.1 基于局部回归的聚类方法

给定样本$x_i$，我们可以根据核函数或者某距离计算方法获得其对应的 $\tau$ 个邻近样本集合$\overline{\boldsymbol{X}}_i=\{\boldsymbol{x}_{i_{(1)}},\boldsymbol{x}_{i_{(2)}},\cdots,\boldsymbol{x}_{i_{(\tau)}}\}$和其对应的聚类结果$\overline{\boldsymbol{Y}}_i=\{\boldsymbol{y}_{i_{(1)}},\boldsymbol{y}_{i_{(2)}},\cdots,\boldsymbol{y}_{i_{(\tau)}}\}$。我们利用这些邻近样本集合构造局部学习模型，用于对样本$x_i$的聚类结果进行预测。本文选择使用非参 Nadaraya-Watson 核回归模型构建核函数，对数据进行非参数估计。与参数估计模型相比，该模型的优势在于：

(1)其复杂性由数据决定，而不是由预先给定的参数决定；

(2)对数据的潜在分布没有要求，适应能力强且稳健性高。

对于样本$\boldsymbol{X}_i$，其邻域集合$\langle\overline{\boldsymbol{X}}_i,\overline{\boldsymbol{Y}}_i\rangle$构建的局部核回归模型对$\boldsymbol{x}_i$的未知聚类结果$\boldsymbol{y}_i$进行预测的结果如式(1)所示：

$$\bar{\boldsymbol{y}}_i=\frac{\sum_{s=1}^{\tau}\boldsymbol{y}_{i_{(s)}}\boldsymbol{K}(\boldsymbol{x}_i,\boldsymbol{x}_{i_{(s)}})}{\sum_{s=1}^{\tau}\boldsymbol{K}(\boldsymbol{x}_i,\boldsymbol{x}_{i_{(s)}})} \tag{1}$$

根据$x_i$的局部回归模型，我们可以定义面向所有样本的局部回归模型的回归系数矩阵 $\boldsymbol{A}\in\mathbb{R}^{n\times n}$，其每个元素$\boldsymbol{A}_{ij}$的定义如式(2)所示：

$$\boldsymbol{A}_{ij}=\begin{cases}\dfrac{\boldsymbol{K}(\boldsymbol{x}_i,\boldsymbol{x}_j)}{\sum_{s=1}^{\tau}\boldsymbol{K}(\boldsymbol{x}_i,\boldsymbol{x}_{i_{(s)}})}, & \text{if } \boldsymbol{x}_j\in\overline{\boldsymbol{X}}_i\\ 0, & \text{if } \boldsymbol{x}_j\notin\overline{\boldsymbol{X}}_i\end{cases} \tag{2}$$

经验证，回归系数矩阵 $\boldsymbol{A}$ 满足两个条件：1) $\boldsymbol{A}_{ij}\geqslant 0$；2) $\sum_{j=1}^{n}\boldsymbol{A}_{ij}=1$。

基于上述核回归模型，我们可以构建基于局部核回归的聚类方法(Multiple Kernel Clustering via Local Regression，CKLR)，其对应的优化问题如式(3)所示：

$$\begin{aligned}&\min\ \|\boldsymbol{Y}-\boldsymbol{A}\boldsymbol{Y}\|^2\\&\text{s.t. }\boldsymbol{Y}^{\mathrm{T}}\boldsymbol{Y}=\boldsymbol{I}\end{aligned} \tag{3}$$

需要指出的是，考虑到局部核回归系数矩阵构造和基于局部核回归的聚类过程，本文提出的 CKLR 方法与传统的局部线性嵌入方法(Local Liner Embedded，LLE)和基于局部学习的聚类方法[17]并不相同，单核意义下的 CKLR 方法具有一定的新颖性。

### 2.2 基于局部回归融合的多核聚类方法

上述基于局部核回归的聚类方法(CKLR)在核空间下较好地考虑了数据的局部流形结构。本节将该方法扩展到多核问题中。

给定 $m$ 个核矩阵$\{\boldsymbol{K}^r\}_{r=1}^m$，我们同样可以构造 $m$ 个局部核回归系数矩阵$\{\boldsymbol{A}^r\}_{r=1}^m$，其定义如式(4)所示：

$$\boldsymbol{A}_{ij}^r=\begin{cases}\dfrac{\boldsymbol{K}^r(\boldsymbol{x}_i,\boldsymbol{x}_j)}{\sum_{s=1}^{\tau}\boldsymbol{K}^r(\boldsymbol{x}_i,\boldsymbol{x}_{i_{(s)}})}, & \text{if } \boldsymbol{x}_j\in\overline{\boldsymbol{X}}_i^r\\ 0, & \text{if } \boldsymbol{x}_j\notin\overline{\boldsymbol{X}}_i^r\end{cases} \tag{4}$$

由式(4)可以看出，针对核矩阵 $\boldsymbol{K}^r$，其核回归邻域 $\overline{\boldsymbol{X}}_i^r$和回归系数都需要根据核矩阵 $\boldsymbol{K}^r$进行适配。

对于多核问题中构造出的多核邻域和多核局部核回归模型，我们可以通过线性加权的方法进一步融合这些回归模型，并得到如下加权的多核局部核回归系数矩阵$\boldsymbol{A}_w\in\mathbb{R}^{n\times n}$，其定义为：$\boldsymbol{A}_w=\sum_{i=1}^{m}\boldsymbol{A}^i$。而且我们同样可以验证回归系数矩阵$\boldsymbol{A}_w$满足以下两个条件：1) $(\boldsymbol{A}_w)_{ij}\geqslant 0$；2) $\sum_{j=1}^{n}(\boldsymbol{A}_w)_{ij}=1$。

因此，我们进一步提出基于局部核回归融合的多核聚类方法(CMKLR)，其对应的优化问题如式(5)所示：

$$\min \| \boldsymbol{Y}-\sum_{i=1}^{m} w_i \boldsymbol{A}^i \boldsymbol{Y} \|^2 \quad \text{s.t.} \ \boldsymbol{Y}^{\mathrm{T}}\boldsymbol{Y}=\boldsymbol{I}, w_i \geqslant 0, \sum_{i=1}^{m} w_i=1 \tag{5}$$

上述优化问题将基于局部核回归聚类和多核局部回归混合系数学习两个子问题进行联合建模。通过这两个任务的联合优化，提高了混合系数学习的质量，进而提高了局部回归精度、稳定性和最终的聚类质量。值得指出的是，本文提出的CMKLR方法一方面借助局部核回归，充分考虑了多核意义下数据的流形结构，有助于提升聚类结果；另一方面待构建的多核局部回归系数矩阵具有较高的稀疏性，减少了多核噪声异常数据的干扰，同样有助于提升聚类结果。

### 2.3　求解算法

本文方法包含两个待求解变量。本文采用常见的坐标轮换法，即固定一组变量求解另一组变量的求解策略。

当变量 $w$ 固定时，定义变量 $\boldsymbol{A}_w=\sum_{i=1}^{m} w_i \boldsymbol{A}^i$，关于变量 $\boldsymbol{Y}$ 的子优化问题为：

$$\min \| \boldsymbol{Y}-\boldsymbol{A}_w \boldsymbol{Y} \|^2 \quad \text{s.t.} \ \boldsymbol{Y}^{\mathrm{T}}\boldsymbol{Y}=\boldsymbol{I} \tag{6}$$

拉普拉斯矩阵的定义如式(7)所示：

$$\boldsymbol{L}_w=(\boldsymbol{I}-\boldsymbol{A}_w)^{\mathrm{T}}(\boldsymbol{I}-\boldsymbol{A}_w)=\boldsymbol{I}-\boldsymbol{A}_w-\boldsymbol{A}_w^{\mathrm{T}}+\boldsymbol{A}_w^{\mathrm{T}}\boldsymbol{A}_w \tag{7}$$

上述优化问题可以进一步简化为：

$$\min \boldsymbol{Y}^{\mathrm{T}}\boldsymbol{L}_w\boldsymbol{Y} \quad \text{s.t.} \ \boldsymbol{Y}^{\mathrm{T}}\boldsymbol{Y}=\boldsymbol{I} \tag{8}$$

由Ky Fan矩阵特征值的定理可知，上述问题的全局最优解 $\boldsymbol{Y}^*$ 对应于拉普拉斯矩阵 $\boldsymbol{L}_w$ 的 $c$ 个最小特征值对应的 $c$ 个特征向量。

当 $\boldsymbol{Y}$ 固定时，定义变量 $\boldsymbol{P}\in \mathbb{R}^{m\times m}$，其中 $\boldsymbol{P}_{ij}=\mathrm{tr}(\boldsymbol{Y}^{\mathrm{T}}\boldsymbol{A}_j^{\mathrm{T}}\boldsymbol{A}_i\boldsymbol{Y})$，定义变量 $\boldsymbol{q}\in \mathbb{R}^{m\times 1}$，其中 $q_i=\mathrm{tr}(\boldsymbol{Y}^{\mathrm{T}}\boldsymbol{A}_i\boldsymbol{Y})$，关于变量 $\boldsymbol{w}$ 的子优化问题为：

$$\min \boldsymbol{w}^{\mathrm{T}}\boldsymbol{P}\boldsymbol{w}-2\boldsymbol{w}^{\mathrm{T}}\boldsymbol{q} \quad \text{s.t.} \ \boldsymbol{w}_i \geqslant 0, \sum_{i=1}^{m} \boldsymbol{w}_i=1 \tag{9}$$

上述带单纯形约束的二次优化问题可以通过现有的二次优化求解器(如Matlab自带的quadprog函数)来求解。

根据上述描述，算法1给出了本文的基于局部回归融合的多核聚类算法。

**算法1　基于局部回归融合的多核聚类算法**

输入：样本间的多个核矩阵 $\{\mathbf{K}^r\}_{i=1}^{m}$，其中 $\mathbf{K}^r\in \mathbb{R}^{n\times n}$，核回归邻域大小 $\tau$，聚类簇数 c

输出：离散化聚类结果 clusterIdx

1. 初始化核函数权重因子 $\mathbf{w}\in \mathbb{R}^{m\times 1}$，$\mathrm{w_i}=1/\mathrm{m}$；
2. 针对每个核矩阵和每个样本，构造其对应的多核回归系数 $\{\mathbf{A}^i\}_{i=1}^{m}$；
3. 计算多核融合的局部回归系数 $\mathbf{A}_w=\sum_{i=1}^{m}\mathrm{w_i}\mathbf{A}^i$；
4. While not converge do.
5. 计算多核局部回归融合的拉普拉斯矩阵 $\mathbf{L}_w=\mathbf{I}-\mathbf{A}_w-\mathbf{A}_w^{\mathrm{T}}+\mathbf{A}_w^{\mathrm{T}}\mathbf{A}_w$；
6. 计算拉普拉斯矩阵 $\mathbf{L}_w$ 的 c 个最小特征值对应的 c 个特征向量 $\mathbf{Y}$；
7. 计算变量 $\mathbf{P}\in \mathbb{R}^{m\times m}$ 和变量 $\mathbf{q}\in \mathbb{R}^{m\times 1}$，其中，$\mathbf{P}_{ij}=\mathrm{tr}(\mathbf{Y}^{\mathrm{T}}\mathbf{A}_j^{\mathrm{T}}\mathbf{A}_i\mathbf{Y})$，$\mathrm{q_i}=\mathrm{tr}(\mathbf{Y}^{\mathrm{T}}\mathbf{A}_i\mathbf{Y})$；
8. 利用 Matlab 中的 quadprog 求解器计算非负多核混合系数 $\mathbf{w}$；
9. 计算当前目标函数 $\mathrm{f_t}$；
10. 如果 $\frac{\mathrm{f_{t-1}}-\mathrm{f_t}}{\mathrm{f_{t-1}}}\leqslant 10^{-5}$，判定收敛，终止迭代；
11. End While.
12. 对 Y 进行归一化，使其每一行的 $\mathrm{L_2}$ 范数为 1 得到 $\hat{\mathbf{Y}}$；
13. 利用 K-means 算法对 $\hat{\mathbf{Y}}$ 进行二次聚类，获得高质量的离散化聚类结果 clusterIdx。

### 2.4　收敛性和复杂性分析

本文提出的基于局部回归融合的多核聚类方法是一个带二次约束和线性约束的二次优化问题。虽然该问题是关于 $\{\boldsymbol{Y},\boldsymbol{w}\}$ 的非凸优化问题且很难获得全局最优解，但是该问题的目标函数明显是有下界的，即下界为0。由于 $\boldsymbol{Y},\boldsymbol{w}$ 求解属于非凸优化问题，因此本文采用分块坐标轮换法进行求解，且都可以获得子问题的最优解。本文算法的每一步求解都使原优化问题的目标函数单调下降，因此本文算法的收敛性得到保证。

本文算法首先需要计算多核回归系数 $\{\boldsymbol{A}^i\}_{i=1}^{m}$，其对应的计算复杂性是 $O(mn\log\tau)$，其中 $m$ 是核函数个数，$n$ 是样本个数，$\tau$ 是局部核回归邻域大小；在迭代过程中构造拉普拉斯矩阵 $\boldsymbol{L}_w$ 和求解特征向量 $\boldsymbol{Y}$ 的复杂度为 $O(n^3)$；在迭代过程中求解多核回归混合权重 $w$ 的复杂度为 $O(m^3)$；收敛后 $K$-means 离散化的复杂度是 $O(nct_2)$，其中 $t_2$ 是 K-means 离散化的迭代次数。综上所述，本文算法的复杂度为 $O(mn\log\tau+(n^3+m^3)t_1+nct_2)$，其中 $t_1$ 是本文算法的迭代次数。考虑到 $m\ll n,\tau\ll n$，以及我们在后续的实验中发现本文算法通常在10次以内就可以收敛，即 $t_1\ll n$，因此本文算法的复杂度可以进一步简化为 $O(n^3)$。

## 3　实验

本节通过在多个不同类型的基准数据集上进行聚类实验，以验证CMKLR算法的有效性。

### 3.1　数据集与预处理

本实验采用5个常见数据集来评估多核聚类方法的性能。这些数据集包括一个生物信息数据Prostate、两个文本数据CSTR和RELATHE、两个图像数据USPS49和UCI-Digit，其中UCIDigit为多视图(6个视图)，其余数据集均为单视图数据。这些数据的相关统计信息如表1所列。

表1　测试数据集的统计信息

Table 1　Statistics of the benchmark data sets

| | 样本个数 | 特征个数 | 类别个数 | 视图个数 | 数据类型 | 多核个数 |
|---|---|---|---|---|---|---|
| Prostate | 102 | 5966 | 2 | 2 | 生物信息 | 12 |
| CSTR | 476 | 1000 | 4 | 4 | 文本 | 12 |
| RELATHE | 1427 | 4322 | 2 | 2 | 图像 | 12 |
| USPS49 | 1673 | 256 | 2 | 2 | 图像 | 12 |
| UCIDigit | 2000 | {6,64,76,47,216,240} | 10 | 10 | 图像 | 12 |

本文采用多核聚类实验中的通用方法来构造多核数据。对于单视图数据，本实验采用7个高斯核函数、4个多项式核函数和1个余弦核函数一共构造12个核矩阵，其中高斯核函数如式(10)所示：

$$k(\boldsymbol{x},\boldsymbol{y})=\exp\left(-\frac{\|\boldsymbol{x}-\boldsymbol{y}\|^2}{2\delta^2}\right) \tag{10}$$

其中，参数 $\delta$ 的取值范围为 $\{0.01D_0, 0.05D_0, 0.1D_0, 1D_0, 10D_0, 50D_0, 100D_0\}$，而 $D_0$ 为样本两两之间的平均欧氏距

离；多项式核函数 $k(x,y)=(a+x^{T}y)^{b}$ 中 $a$ 的取值包括{0，1}，$b$ 的取值包括{2，4}；余弦函数为 $k(x,y)=\frac{x^{T}y}{\|x\|\cdot\|y\|}$。对于多视图数据 UCIDigit，我们采用高斯核函数和余弦核函数为每个视图构造两个核矩阵，其中高斯核的超参数采用样本间的平均欧氏距离。

### 3.2 对比算法和参数设置

为了验证本文提出的多核局部回归聚类方法（CMKLR）的有效性，本实验在 5 个基准数据集上与以下 10 种主流多核聚类算法进行对比，其中包括基于协同训练的多视图谱聚类方法（CTSC）[5]、基于协同正则的多视图谱聚类方法（Coreg）[6]、鲁棒多视图谱聚类方法（RMSC）[7]、鲁棒多核 $K$-means 方法（RMKKM）[4]、基于矩阵正则的多核 $K$-means 方法（MKKMMR）[8]、基于多核局部对齐的方法（LKAMKC）[9]、基于最优邻近核学习的多核聚类方法（ONMKC）[10]、基于低秩学习的多核聚类方法（LKGr）[11]、基于自适应鲁棒核子空间学习的聚类方法（JMKSC）[12]、基于一致性结构化图学习的多视图聚类方法（MCGC）[13]。

其中 MKKMMR，LKAMKC，ONMKC 和 JMKSC 这 4 种方法的源码由文献[4，9-12]提供，其余 6 种方法的源码均可以在相关网站上获得。即对于上述 10 种对比方法，我们均采用论文作者提供的源码进行对比。

需要指出的是，这些聚类方法往往涉及超参数的搜索和设置等。为了提高论文实验结果的可靠性和可重复性，我们采用与其他论文[12]类似的设置，具体如下：

(1)在聚类过程中我们使用真实类别个数作为聚类簇的个数。

(2)超参数设置始终是聚类方法面临的挑战之一，本实验采用通用的网格化搜索策略，即展示不同超参数组合下某方法达到的最佳性能。

(3)CTSC，Coreg 和 RMSC 都采用默认参数。

(4)RMKKM 的超参数搜索范围为{0.1，0.2，…，0.9}，MKKMMR 的超参数搜索范围为{$2^{-15}$，$2^{-14}$，…，$2^{15}$}，LKAMKC 的两个超参数搜索范围为{0.1，0.2，…，1}和{$2^{-15}$，$2^{-14}$，…，$2^{15}$}，ONMKC 的两个超参数搜索范围均为{$2^{-15}$，$2^{-13}$，…，$2^{15}$}，JMKSC 的 3 个超参数搜索范围为{$10^{-4}$，$10^{-3}$，…，$10^{1}$}，{1，5，10，15，20，25，30}和{0.1，1，5，10，15，20，25，30}，MCGC 邻域超参数的搜索范围为{3，5，7，9，11，13，15}；这些超参数的具体解释可以参考文献[4，8-13]。本文提出的 CMKLR 算法超参数，即邻域 $\tau$ 的搜索范围为{3，5，7，9，11，13，15}。从候选搜索参数的角度来看，本文提出的 CMKLR 算法仅仅拥有一个超参数 $\tau$。

(5)上述 11 种方法中除 RMKKM 和 MCGC 外，都需要进行后处理以获得离散化的聚类结果，本实验采用 $K$-means 进行离散化。需要指出的是，$K$-means 的结果受初始化的影响较大。为了减小随机初始化的影响，本实验对每种方法和每个超参数组合均运行 20 次 $K$-means 算法，最终记录 20 次 $K$-means 运行中获得的 $K$-means 算法最小目标函数值对应的聚类结果。

(6)本实验采用常用的 3 个指标来评估算法性能，其中包括聚类精度（Clustering Accuracy，ACC）、归一化互信息（Normalized Mutual Information，NMI）和聚类纯度（clustering purity）[12]。

通过上述设置，我们构建了较为公平的测试环境，以便准确评估这些算法在给定数据集上的聚类性能。

为了提高实验的可重复性，本文中的相关代码已开源至相关网站[1)]。

### 3.3 多核聚类实验结果

表 2－表 4 列出了这 11 种算法在 5 个数据集上对应的聚类精度、归一化互信息和聚类纯度 3 种指标的结果。各评价指标的简介如下。

(1)聚类精度（ACC）。聚类准确性[18]指标揭示了聚类后得到的类簇信息和真实类别的对应关系，以对聚类方法的性能给出客观评价。给定一个点 $x_i$，$p_i$ 和 $q_i$ 分别代表聚类的结果以及真实标签，则 ACC 的定义如式(11)所示：

$$ACC=\frac{1}{n}\sum_{i=1}^{n}\delta(q_i,map(p_i)) \tag{11}$$

其中，$n$ 代表样本的总数，$\delta(x,y)$ 指 $\delta$ 函数。如果 $x=y$，则 $\delta(x,y)=1$；如果 $x\neq y$，则 $\delta(x,y)=0$。map(·)代表置换映射函数，它将每个簇索引映射到一个真实的类标签。实验中一般采用 Kuhn-Munkres 算法找到最佳映射。$ACC$ 取值为 0～1，一般情况下，聚类准确性越高，聚类的性能就越好。

(2)归一化互信息（NMI）。归一化互信息[19]是聚类问题中常用的外部评价指标，用于确定聚类质量，评价算法在一个数据集上的聚类结果与该数据集真实划分的相似程度。设 $C$ 表示真实标签中类簇的集合，$C'$ 表示从聚类算法得到的类簇的集合，其相互信息 $MI(C,C')$ 的定义如式(12)所示：

$$MI(C,C')=\sum_{c_i\in C,c_j'\in C'}p(c_i,c_j')\log\frac{p(c_i,c_j')}{p(c_i)p(c_j')} \tag{12}$$

其中，$p(c_i)$ 和 $p(c_j')$ 分别代表从数据集中任意选择的数据点属于簇 $c_i$ 和 $c_j'$ 的概率，$p(c_i,c_j')$ 代表任意选择的数据点同时属于簇 $c_i$ 和 $c_j'$ 的联合概率。在本文的实验中，我们统一采用归一化的互信息，其定义如式(13)所示：

$$NMI(C,C')=\frac{MI(C,C')}{\max(H(C),H(C'))} \tag{13}$$

其中，$H(C)$ 和 $H(C')$ 分别代表 $C$ 和 $C'$ 对应的信息熵。同样，$NMI$ 取值为 0～1，$NMI$ 越高，聚类的性能就越好。

(3)聚类纯度（clustering purity）。该指标用于测量正确聚类的样本数占样本总数的比例。聚类纯度可以由单个聚类纯度值的加权得到，其具体定义如式(14)所示：

$$Purity=\frac{1}{n}\sum_{k}\max(c_k',c_j) \tag{14}$$

其中，$C=\{c_1,c_2,\cdots,c_k\}$ 表示真实标签中类的集合；$C'=\{c_1',c_2',\cdots,c_k'\}$ 表示聚类算法获得的类簇集合。$Purity$ 的取值也为 0～1，$Purity$ 值越高，聚类的性能就越好。

值得指出的是，除 RMKKM 和 MCGC 算法外，本实验汇报的是基于 20 次 $K$-means 随机初始化对应最小 $K$-means 目标函数值的离散化结果；实验中 RMKKM 采用 20 次随机初始化对应最小 RMKKM 目标函数值的离散化结果；而 MCGC 则是通过连通图分割的方法来获取离散化结构，不受随机初始化的影响。因此，在这 3 个实验结果表中没有方差等其他

1) https://gitee.com/csliangdu/CMKLR

统计信息。此外，本文提出的 CMKLR 算法的结果和最佳结果都用加粗的数字表示。

表 2　11 种聚类方法在 5 个测试数据集上的聚类精度结果

Table 2　Clustering accuracy of 11 compared algorithms on 5 data sets

| | Prostate | CSTR | RELATHE | USPS49 | UCIDigit | Average |
|---|---|---|---|---|---|---|
| CTSR | 0.5784 | 0.6092 | 0.6412 | 0.7095 | 0.9000 | 0.6877 |
| Coreg | 0.5784 | 0.5924 | 0.6741 | 0.7400 | 0.3185 | 0.5807 |
| RMSC | 0.5784 | 0.5798 | 0.5809 | 0.7466 | 0.8705 | 0.6712 |
| RMKKM | 0.5784 | 0.6261 | 0.5683 | 0.7854 | 0.4875 | 0.6091 |
| MKKMMR | 0.6176 | 0.5903 | 0.6300 | 0.7657 | 0.8905 | 0.6988 |
| LKAMKC | 0.6176 | 0.7185 | 0.7169 | 0.7651 | 0.9355 | 0.7507 |
| ONMKC | 0.6176 | 0.5945 | 0.7281 | 0.7675 | 0.9160 | 0.7247 |
| LKGr | 0.6275 | 0.7920 | 0.5830 | 0.7884 | 0.9205 | 0.7423 |
| JMKSC | 0.5980 | 0.6639 | 0.5718 | 0.8452 | 0.9360 | 0.7230 |
| MCGC | 0.6373 | 0.5882 | 0.6601 | 0.7310 | 0.9455 | 0.7124 |
| CMKLR | **0.6765** | **0.7773** | **0.8297** | **0.9074** | **0.9645** | **0.8311** |

表 3　11 种聚类方法在 5 个测试数据集上的归一化互信息结果

Table 3　Clustering NMI of 11 compared algorithms on 5 data sets

| | Prostate | CSTR | RELATHE | USPS49 | UCIDigit | Average |
|---|---|---|---|---|---|---|
| CTSR | 0.0178 | 0.5624 | 0.0990 | 0.1328 | 0.8174 | 0.3259 |
| Coreg | 0.0178 | 0.5489 | 0.1272 | 0.1731 | 0.1849 | 0.2104 |
| RMSC | 0.0176 | 0.5302 | 0.0548 | 0.1832 | 0.7921 | 0.3156 |
| RMKKM | 0.0178 | 0.5485 | 0.0090 | 0.2532 | 0.5147 | 0.2686 |
| MKKMMR | 0.0426 | 0.5417 | 0.0908 | 0.2165 | 0.8143 | 0.3412 |
| LKAMKC | 0.0492 | 0.5876 | 0.1537 | 0.2147 | 0.8747 | 0.3760 |
| ONMKC | 0.0426 | 0.5508 | 0.1724 | 0.2219 | 0.8541 | 0.3684 |
| LKGr | 0.0574 | 0.6129 | 0.0379 | 0.2567 | 0.8558 | 0.3641 |
| JMKSC | 0.0401 | 0.5118 | 0.0295 | 0.3783 | 0.8866 | 0.3693 |
| MCGC | 0.0592 | 0.4954 | 0.1134 | 0.1602 | 0.8866 | 0.3430 |
| CMKLR | **0.0964** | **0.6364** | **0.3752** | **0.5783** | **0.9198** | **0.5212** |

表 4　11 种聚类方法在 5 个测试数据集上的聚类纯度结果

Table 4　Clustering purity of 11 compared algorithms on 5 data sets

| | Prostate | CSTR | RELATHE | USPS49 | UCIDigit | Average |
|---|---|---|---|---|---|---|
| CTSR | 0.5784 | 0.7605 | 0.6412 | 0.7095 | 0.9000 | 0.7179 |
| Coreg | 0.5784 | 0.7542 | 0.6741 | 0.7400 | 0.3265 | 0.6146 |
| RMSC | 0.5784 | 0.7437 | 0.5809 | 0.7466 | 0.8705 | 0.7040 |
| RMKKM | 0.5784 | 0.7542 | 0.5683 | 0.7854 | 0.5305 | 0.6434 |
| MKKMMR | 0.6176 | 0.7542 | 0.6300 | 0.7657 | 0.8905 | 0.7316 |
| LKAMKC | 0.6176 | 0.7836 | 0.7169 | 0.7651 | 0.9355 | 0.7637 |
| ONMKC | 0.6176 | 0.7563 | 0.7281 | 0.7675 | 0.9160 | 0.7571 |
| LKGr | 0.6275 | 0.7920 | 0.5830 | 0.7884 | 0.9205 | 0.7423 |
| JMKSC | 0.5980 | 0.7500 | 0.5718 | 0.8452 | 0.9360 | 0.7402 |
| MCGC | 0.6373 | 0.7311 | 0.6601 | 0.7310 | 0.9455 | 0.7410 |
| GMKLR | **0.6765** | **0.7920** | **0.8297** | **0.9074** | **0.9645** | **0.8340** |

我们可以非常容易地从表 2－表 4 的实验结果中看出，在所有数据集中，本文提出的聚类算法 CMKLR 几乎在所有指标上都优于其他强有力的对比算法。具体来看，CMKLR 在 5 个数据集中的聚类精度平均达到了 0.8311，高出第二名 LAMKC 算法 10.71%；CMKLR 在 5 个数据集中的归一化互信息平均达到了 0.5212，高出第二名 LKAMKC 算法 38.62%；而 CMKLR 在 5 个数据集中的聚类纯度平均达到了 0.8340，高出第二名 LKAMKC 算法 9.21%。

更进一步，可以看出最近提出的算法如 LKAMKC，ONMKC，LKGr，JMKSC 和 MCGC 整体上优于 CTSR，Coreg，RMSC，RMKKM 和 MKKMMR。另外，CMKLR，LKAMKC 这两种局部学习算法的表现均优于其他全局类算法，如 ONMKC，LKRr 和 JMKSC，可能的原因之一是这两种算法都在多核数据的聚类问题中显式考虑了数据间的流形结构。尽管局部流形结构的重要性已经在多个无监督学习场景中得到充分验证，但是现有多核聚类方法较少关注多核意义下的局部流形结构。尽管 MCGC 也在通过自适应学习的方法学习拥有分块对角结构的一致性相似图，但是在多核数据上的表现并不是特别好。这里可能的原因之一是 MCGC 方法[13]针对原始数据通过 PCAN 方法构造单视图上的初始化图，但是在本文的多核实验中采用从核矩阵中抽取 $\tau$ 邻近图作为单视图初始化图，而低质量的核矩阵往往也会导致低质量的 $\tau$ 邻近图，进而降低了输入数据的质量，从而使得 MCGC 算法在多核聚类问题上的性能下降。

上述观测和分析表明，本文提出的 CMKLR 算法在多核聚类问题上优于许多主流聚类方法。

### 3.4　收敛性实验结果

本文以 CSTR 数据集为例，给出了本文算法的收敛情况，如图 1 所示，首先目标函数值在迭代过程中呈单调递减的趋势，其次目标函数在迭代较少次数后(10 次)出现收敛。

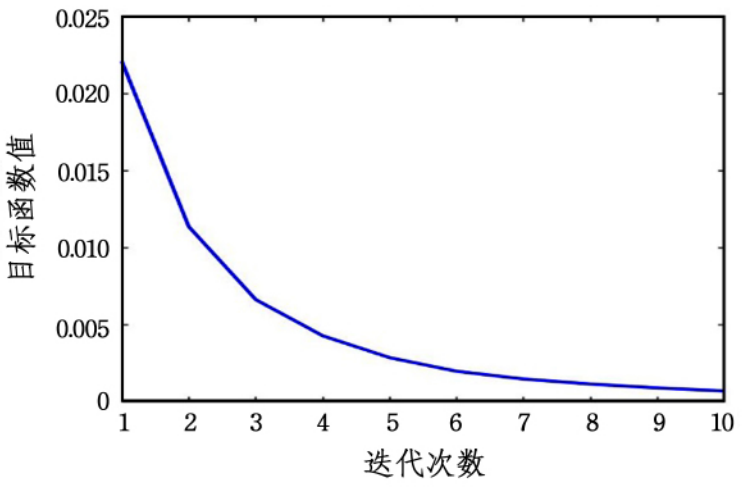

**图 1 收敛性趋势图**

Fig. 1 Convergence trend diagram

**结束语** 本文提出基于局部回归融合的多核聚类方法。该方法通过单核局部回归融合的方式很好地刻画了多核数据的局部流形结构，同时通过加权融合单核局部回归系数的方式得到稀疏的多核局部回归系数，极大地缩减了待估计变量的规模，减少了多核噪声异常的干扰。本文设计了对应的分块迭代求解算法，并证明了其收敛性和复杂性。基准数据集上的多核聚类实验结果表明，本文方法优于当前主流的多核聚类方法。下一步的研究将主要集中在如何减轻唯一的超参数即局部回归邻域大小对算法的影响。

## 参考文献

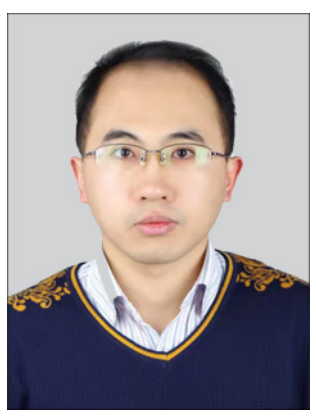

**DU Liang**, born in 1985, Ph.D, is a member of China Computer Federation. His main research interests include machine learning and data mining.

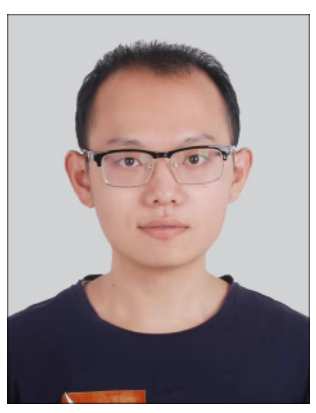

**ZHOU Peng**, born in 1989, Ph.D, is a member of China Computer Federation. His main research interests include data mining and machine learning.